\documentclass[letterpaper, 10 pt, conference]{ieeeconf}  

\IEEEoverridecommandlockouts                              

\title{\LARGE \bf
Hybrid Physical Metric For 6-DoF Grasp Pose Detection
}

\author{Yuhao Lu, Beixing Deng, Zhenyu Wang, Peiyuan Zhi, Yali Li, Shengjin Wang\\
Tsinghua University
}

\usepackage[T1]{fontenc}
\usepackage{cite}
\usepackage{graphicx}
\usepackage{booktabs}
\usepackage{multirow}
\usepackage{multicol}
\usepackage{amsmath}
\usepackage{amsfonts}
\usepackage{cleveref}
\usepackage{tablefootnote}

\newcommand{\myref}[1]{Eq.\ref{#1}}
\newcommand{\figref}[1]{Fig.~\ref{#1}}

\begin{document}

\maketitle
\thispagestyle{empty}
\pagestyle{empty}

\begin{abstract}
6-DoF grasp pose detection of multi-grasp and multi-object is a challenge task in the field of intelligent robot. To imitate human reasoning ability for grasping objects, data driven methods are widely studied. With the introduction of large-scale datasets, we discover that a single physical metric usually generates several discrete levels of grasp confidence scores, which cannot finely distinguish millions of grasp poses and leads to inaccurate prediction results. In this paper, we propose a hybrid physical metric to solve this evaluation insufficiency. First, we define a novel metric is based on the force-closure metric, supplemented by the measurement of the object flatness, gravity and collision. Second, we leverage this hybrid physical metric to generate elaborate confidence scores. Third, to learn the new confidence scores effectively, we design a multi-resolution network called Flatness Gravity Collision GraspNet (FGC-GraspNet). FGC-GraspNet proposes a multi-resolution features learning architecture for multiple tasks and introduces a new joint loss function that enhances the average precision of the grasp detection. The network evaluation and adequate real robot experiments demonstrate the effectiveness of our hybrid physical metric and FGC-GraspNet. Our method achieves 90.5\% success rate in real-world cluttered scenes. Our code is available at https://github.com/luyh20/FGC-GraspNet.

\end{abstract}

\section{INTRODUCTION}

Grasping is one of the most fundamental and important tasks in the field of robotic manipulation. Recently data-driven methods \cite{kumra2017robotic, redmon2015real, ten2017grasp} have been developed to reason robust grasps under various settings. For the point cloud data of observed scenes, the model outputs grasp poses by a deep neural network \cite{fang2020graspnet, zhao2020regnet, sundermeyer2021contact}. However, generating reliable and humanoid grasps for multiple objects in unstructured and cluttered environments is still a challenge. 

\begin{figure}[t]
    \centering
    \includegraphics[width=\linewidth]{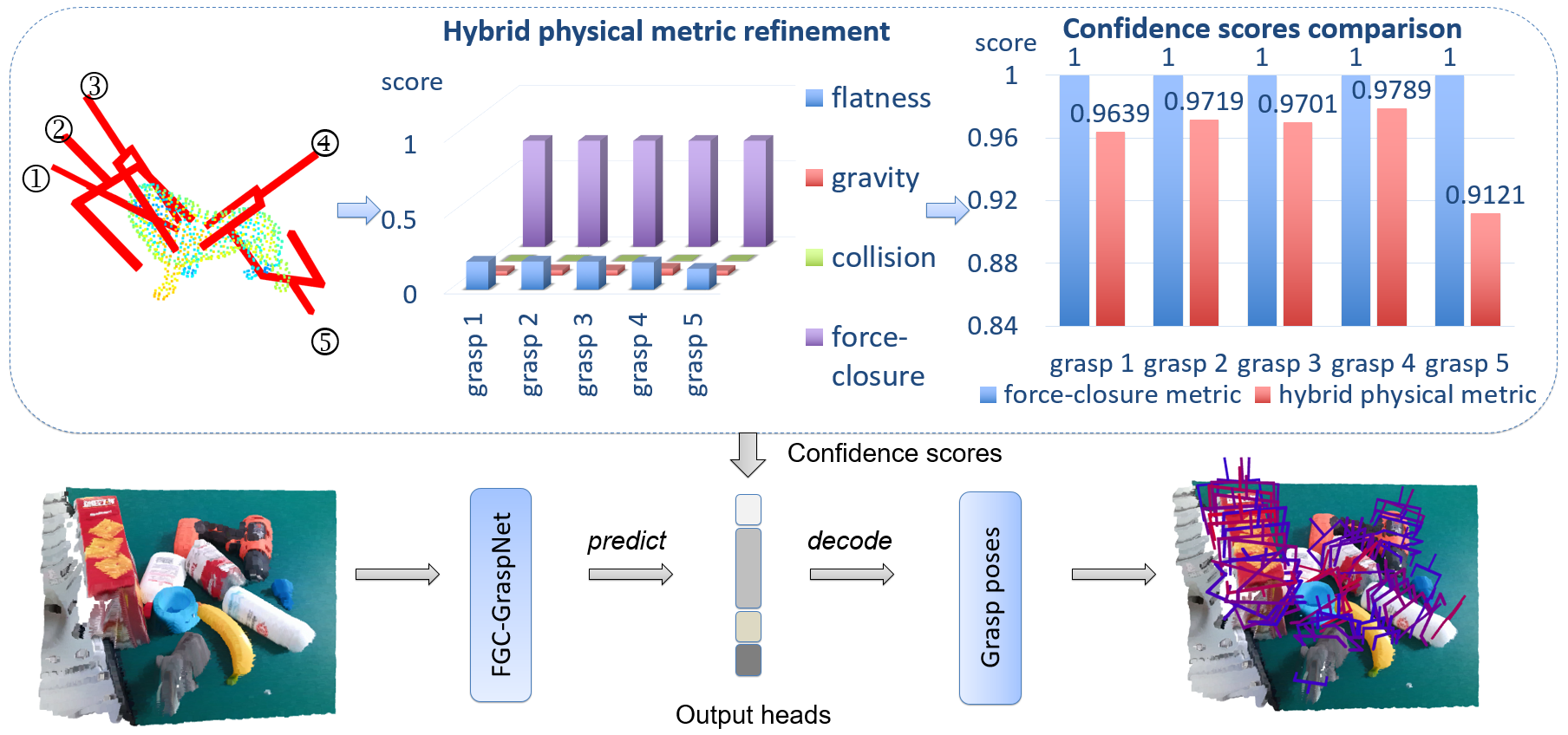}
    \caption{Top: Grasp confidence scores refinement. There are five grasp candidates labeled in the lion model. The middle chart shows the composition of hybrid physical metric. The right chart is the comparison results. Bottom: Grasp pose detection pipeline. The input data forwards through our FGC-GraspNet and predicts grasp poses.}
    \label{fig:first}
\end{figure}

The performance and efficacy of data-driven methods mainly depends on two aspects: inferring the grasp quality and predicting the grasp pose. Since it is laborious and costly to annotate the potential success rates of large-scale 6-DoF grasp poses by human, a solid grasp evaluation mechanism to infer the grasp quality is of vital importance. Some recent grasp pose detection methods\cite{kumra2017robotic, ten2017grasp, liang2019pointnetgpd, zhao2020regnet, fang2020graspnet} apply the physics analytic approaches to evaluate the quality of grasp poses. Thereinto, the force-closure metric\cite{tung1996fast} is a mainstream evaluation metric. Although it has been utilized by many works \cite{liang2019pointnetgpd, fang2020graspnet}, it is still limited. The main problem is that it only provides a binary outcome under a coefficient of friction, and usually generates confidence scores in several ranked discrete levels, like ten bins in \cite{fang2020graspnet}. Such discrete score levels necessarily exist noise. As is illustrated in the top of \figref{fig:first}, some grasp poses apparently achieve different grasp performances but are assigned in the same level score under force-closure metric. Because of the noise, the network is likely to predict confidence scores deviated from the truth. Such a confidence score gap between real world grasp success rates and calculated confidence scores affects the adaptability of the network in real world application. In addition, the force-closure metric only describes a single physical characteristic, which is quite insufficient to reason reliable grasp poses when encountering novel objects and degrades the robustness of grasp pose detection.

To tackle these issues above, in this paper, we propose a hybrid physical metric to solve the evaluation insufficiency. To reduce the influence of label noise, we leverage more comprehensive physical information to refine grasp confidence scores. According to human grasp habits, we pay attention to the contact points between the two-finger gripper and the corresponding object. The first evaluation metric we adopt is the flatness metric, which aims to measure the flatness of contact points. The second is the gravity center metric, and the motivation is to balance the gravity and the grasping pressure. The third is the collision perturbation metric, which prevents possible collisions between grasp end points and contact points. Together with force-closure metric, we present the hybrid physical metric to evaluate the final grasp confidence scores. As is shown in the top part of \figref{fig:first}, our designed hybrid physical metric is used to distinguish grasp candidates more reasonably than force-closure metric.

Meanwhile, to learn this fine hybrid metric more effectively, we further design a multi-resolution network called FGC-GraspNet. During the training process, the grasp confidence scores are passed into different loss functions for the multi-task learning paradigm \cite{ruder2017overview, sener2018multi}, such as the approaching direction or the in-plane rotation prediction \cite{fang2020graspnet}. However, different tasks usually require different kinds of information. Specifically, the predecessor tasks, including foreground segmentation and approaching vector prediction, often need scene-level and object-level information, while the post tasks like in-plane rotation and depth prediction focus more on object-level and point-level information. Hence, FGC-GraspNet is a multi-resolution network. Based on the hierarchical feature learning of PointNet++\cite{qi2017pointnet++}, the features of low-resolution point sets are extracted for the predecessor tasks. The features of high-resolution point sets forward a local attention module to gather region information for the post tasks. Furthermore, to better accommodate to our hybrid metric, we adopt a new joint learning loss function. This loss regresses the confidence scores more sufficiently and contains a grasp depth classification loss.

We implement the hybrid physical metric on the GraspNet-1Billion dataset\cite{fang2020graspnet}, and obtain more elaborate and physically meaningful grasp confidence scores. We then use the new grasp confidence scores in our grasp pose detection work. Experiment results demonstrate that our FGC-GraspNet under new grasp confidence scores achieves significantly better performance.

To summarize, our main contributions are as follows:
\begin{itemize}

\item We propose a hybrid physical metric to refine the grasp confidence scores. This metric adopts more comprehensive physical descriptions thus is more reasonable.

\item We design a multi-resolution network FGC-GraspNet. By utilizing information of different levels of resolution for multiple tasks and adopting new joint loss function, our network better adapts to our hybrid physical metric.

\item Extensive experiments show that the hybrid physical metric is beneficial for increasing the success rates in reality, and our network significantly promotes the ability of grasp prediction.

\end{itemize}

\section{RELATED WORK}

\subsection{6-DoF Grasp Pose Detection}
Most of recent 6-DoF grasp pose detection works are based on data-driven methods, where 6-DoF is decoupled to 3D position and 3D rotation vector for the movement of the robotic arm. Compared to rectangle grasp representation\cite{jiang2011efficient, goldfeder2009columbia}, 6-DoF allows multi-view camera inputs and provides more possible approaching directions of grasp poses. 
Many recent methods\cite{varley2017shape, ten2017grasp, mousavian20196, gou2021rgb, mahler2017dex, zhao2020regnet, qin2020s4g} 
attempt to learn predicting grasp poses based on deep learning. PointNetGPD\cite{liang2019pointnetgpd} samples grasp candidates and evaluates the grasp quality based on the network PointNet\cite{qi2017pointnet}. GraspNet-1Billion\cite{fang2020graspnet} builds a large-scale grasp dataset and proposes a baseline method for learning grasp poses. REGNet\cite{zhao2020regnet} use group region features to predict grasp proposals. Contact-GraspNet\cite{sundermeyer2021contact} proposes a new grasp pose representation and implement it in \cite{eppner2020acronym}. A object instance segmentation network\cite{xie2020best} is utilized in \cite{murali20206} to tackle the problem of grasping objects in a cluttered scene. These methods greatly enrich the solutions of grasp pose detection. However, most of them evaluate the grasp quality at discrete levels. Therefore, we propose hybrid physical metric and design a multi-resolution network to learn the refined confidence scores.

\begin{figure}[t]
\centering
\begin{minipage}{0.8\textwidth}
\includegraphics[width=0.5\linewidth]{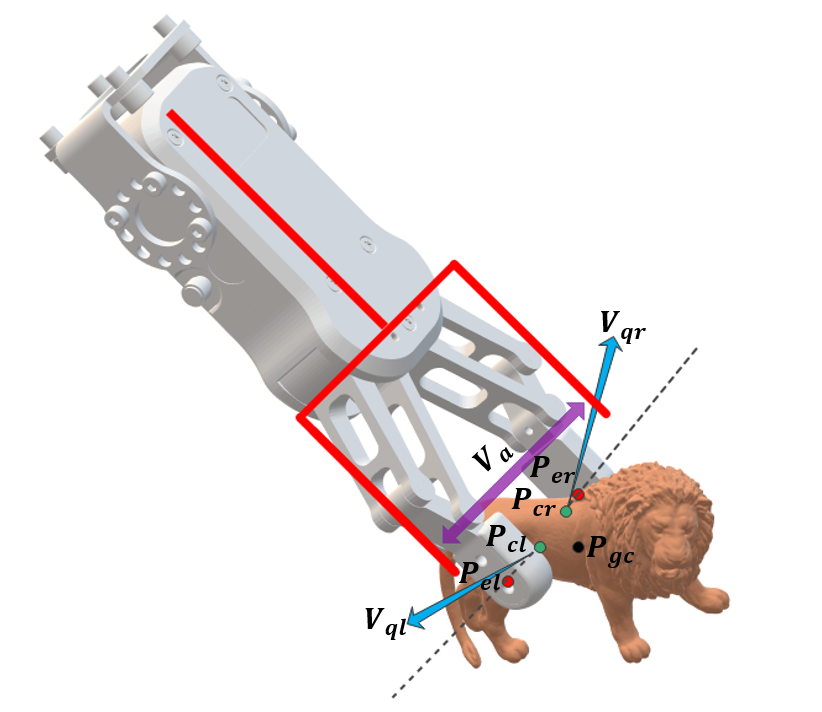}
\end{minipage}
\caption{Grasp pose demonstration. $P_{cl}$, $P_{cr}$ are the contact points in the object, $P_{el}$, $P_{er}$ are the end points of gripper, $P_{gc}$ is the gravity center of the object, $V_a$ is the antipodal vector, $V_{ql}$, $V_{qr}$ are the normals of the contact points. The dotted line is the reference line of $V_n$. The gripper model is from RG2 \protect\footnotemark[1]. The lion model is from the dataset\cite{fang2020graspnet}. }
\label{fig:funny}
\end{figure}

\subsection{Grasp Evaluation Metrics}
Grasp quality evaluation is essential component of supervised learning grasp detection methods. A reliable grasp labels should reflect the grasping success rate in real robotic manipulation accurately. A recent work like\cite{levine2018learning} produces grasp labels through a large-scale real robotic grasping attempts. Some works\cite{kappler2015leveraging, depierre2018jacquard, yan2018learning, eppner2020acronym, mahler2017dex} utilize physics simulators to annotate grasp confidence scores. Many analysis approaches\cite{kumra2017robotic, ten2017grasp, liang2019pointnetgpd, zhao2020regnet, fang2020graspnet} generate confidence scores by calculating the contact physics about the geometry of gripper configuration and object mesh models. Defining a reasonable grasp quality metric is an open problem. Fingertip space is proposed in \cite{6942775} to search stable grasps by considering both the local geometry of object surface and the fingertip geometry and a flatness criteria is defined to filter points.  PointNetGPD\cite{liang2019pointnetgpd} modify  the coefficient of friction to get a discrete score based on force-closure\cite{nguyen1988constructing} and use grasp wrench space (GWS)\cite{kirkpatrick1992quantitative} analysis as a auxiliary. GraspNet-1Billion\cite{fang2020graspnet} also calculate friction coefficient of grasp poses at 10 bins. REGNet\cite{zhao2020regnet} measure the angle between the force direction and contacts’ normals. Four different grasp stability metrics are applied in \cite{kappler2015leveraging}, including $\epsilon-metric$\cite{pokorny2013classical} and so on. These metrics have respective limitations for 6-DoF grasp pose detection. Hence, we propose the hybrid physical metric in this paper.

\subsection{Point Cloud Learning Methods}
Point cloud learning is one of necessary module during grasp pose detection methods of data driven. A variety of approaches are proposed for the features learning of point cloud, consisting of point-wise MLP methods such as PointNet\cite{qi2017pointnet}, PointNet++\cite{qi2017pointnet++},  convolution-based  methods such as PointConv\cite{wu2019pointconv}, PointCNN\cite{li2018pointcnn}, and some newly proposed transformer-based methods such as PCT\cite{guo2020pct}, PT\cite{zhao2020point}. In the field of 3D object detection, Local-Global transformer module\cite{pan20213d} and a two-stage architecture for multi-task learning\cite{shi2019pointrcnn} are proved useful in generating 3D object proposals. Currently, PointNet++\cite{qi2017pointnet++} is used widely in the area of grasp pose detection.

\begin{figure}[t]
\centering
\begin{minipage}{0.9\textwidth}
\includegraphics[width=0.5\linewidth]{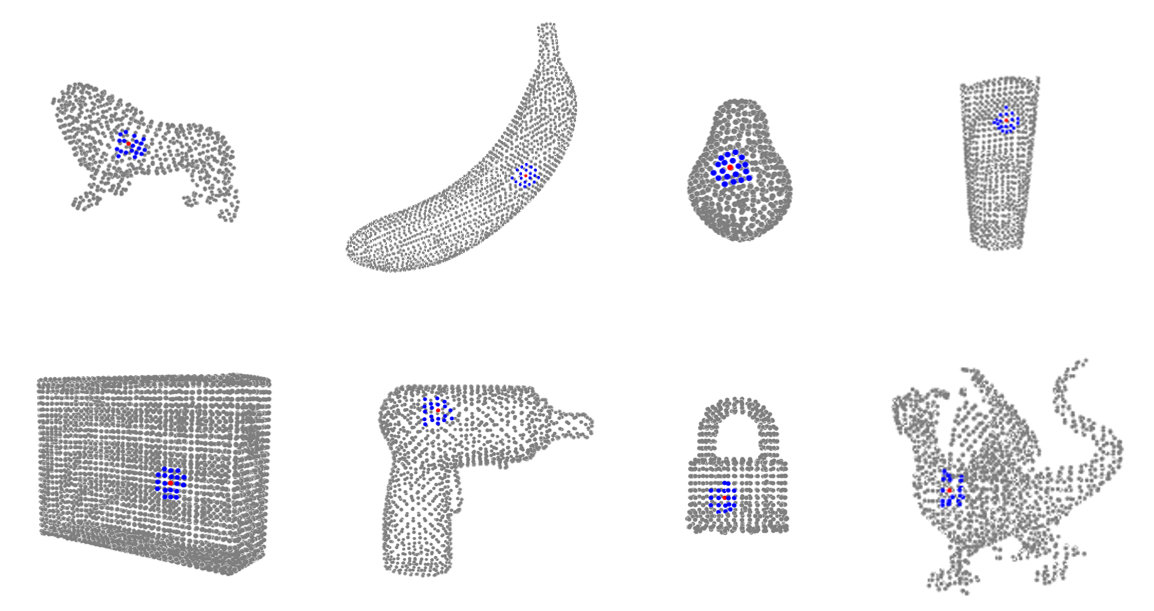}
\end{minipage}
\caption{Flatness measurement of partial object mesh models. These object models are from the dataset\cite{fang2020graspnet}. The red point is the flattest point, indicating the highest $S_f$ among all points. The blue points are the neighbors points.}
\label{fig:flatness}
\end{figure}

\section{HYBRID PHYSICAL METRIC}

In this section, we introduce our proposed hybrid physical metric. Given the 3D object mesh model and the grasp pose annotated in the model, the grasp evaluation process aims to predict the confidence scores for different grasp poses. We use the same 6-DoF grasp representation to define the grasp pose with the dataset\cite{fang2020graspnet}. Based on the previous force-closure metric, we further adopt flatness metric, gravity center metric and collision perturbation metric, to generate more accurate grasp confidence scores on the GraspNet 1-Billion dataset\cite{fang2020graspnet}. Specifically, we focus on the contact points, the critical geometric points and vectors, which are displayed in \figref{fig:funny}.



\textbf{Flatness Metric}.
For the grasp action of a two-finger gripper, intuitively, the grasp quality is higher when the contact region is more flat. To utilize this information, we quantify the flatness of the contact region in two steps.

First, we calculate the flatness of points in 3D mesh models. The similarity of the local normal vectors in the query point region can be used to measure the flatness score of the points, denoted with $S_{f1}$. We use cosine distance between k-nearest neighbors' normals and the query point normal to calculate it. Partial flatness measurement results with our designed score are shown in \figref{fig:flatness}. Second, we consider the perpendicularity between the antipodal direction and the contact region. We calculate the consistency between the antipodal vector and the contact point normal with the cosine distance as the consistency score $S_{f2}$. The final score $S_{f}$ is obtained through the multiplication operation of these two scores. The specific calculation operation can be formulated as follows:

\begin{equation}
\begin{split}\label{eq:1}
    &S_{f1} = \frac{1}{2K}\sum_{q=1}^{2} \sum_{n=1}^{K} \frac{<V_q, V_n > }{||V_q||\cdot||V_n||}  \\
 &S_{f2} = \frac{1}{2}\sum_{q=1}^{2} \frac{|<V_a, V_q >| }{||V_a||\cdot||V_q||}  \\
 &S_{f} = S_{f1} \cdot S_{f2}
\end{split}
\end{equation}
where $V_q$ is the normal vector of the query point, including two contact points, $V_n$ is the normal vector of its neighbor points, $K$ is the number of neighbors and $V_a$ is the antipodal vector. $<\cdot, \cdot>$ denotes the inner product operation.

\footnotetext[1]{$https://github.com/ekorudiawan/rg2\_simulation$}

\textbf{Gravity Center Metric}.
Considering the fact that the grasp candidate whose antipodal force is closer to the gravity center of object is more steady, we propose the gravity center metric. We thus quantify this mechanical relationship between the gravity and the pressure into a distance metric. In the light of the geometry of the grasp pose, the two contact points are linked to an antipodal line. We adopt the euclidean distance between the gravity center point and the antipodal line as the gravity score $S_g$:
\begin{equation}\label{eq:2}
    S_g = \frac{||(P_{cl}-P_{gc})\times(P_{cr}-P_{gc})||}{||P_{cl}-P_{cr}||}
\end{equation}
where $P_{cl}$, $P_{cr}$ and $P_{gc}$ are coordinates of the left contact point, right contact point and the gravity center point. The $S_g$ is normalized and converted  to $1-S_g$ in practice.

\begin{figure*}[t]
\centering
\begin{minipage}{0.9\textwidth}
\center{\includegraphics[width=\linewidth]{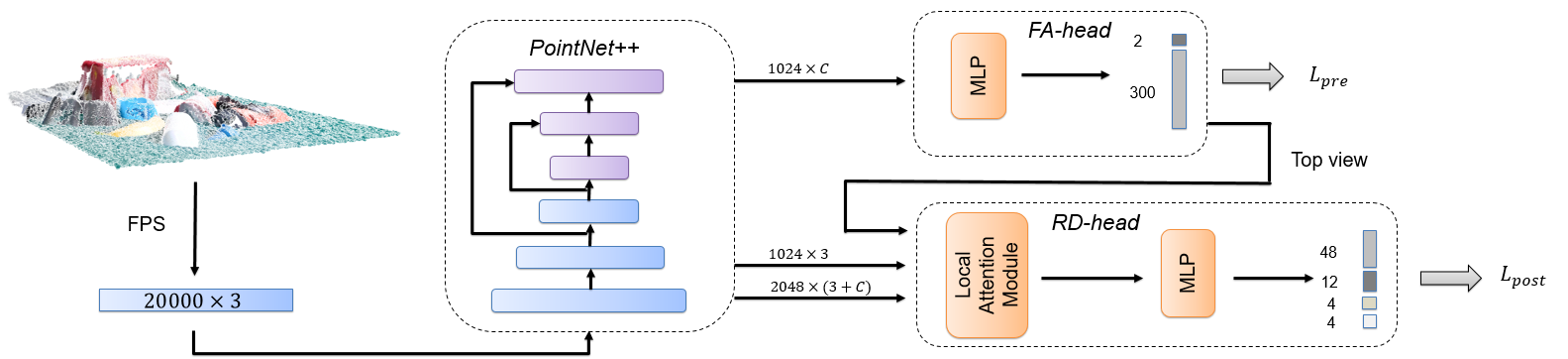}}
\end{minipage}
\caption{The architecture of FGC-GraspNet. The input point clouds are sampled by farthest point sampling (FPS)\cite{qi2017pointnet++} to $20000\times 3$. The network consists of PointNet++\cite{qi2017pointnet++}, FA-head(Foreground-Approach-head), RD-head(Rotation-Depth-head), and more details in text.}
\label{fig:net}
\end{figure*}

\textbf{Collision Perturbation Metric}.
We observe that the grasp candidates are prone to collision when the end point is close to the object model in real world experiments. Hence, to avoid too close contact, the minimum value of the euclidean distances between the end points and the object contact points is formulated as the collision perturbation score $S_{c}$, 
\begin{equation}
    S_{c} = \min(||P_{el}-P_{cl}||, ||P_{er}-P_{cr}||)
\end{equation}
where $P_{el}$, $P_{er}$ are coordinates of the left end point and the right end point. The $S_c$ is also normalized.

\begin{figure*}[t]
\centering
\begin{minipage}{0.9\textwidth}
\center{\includegraphics[width=\linewidth]{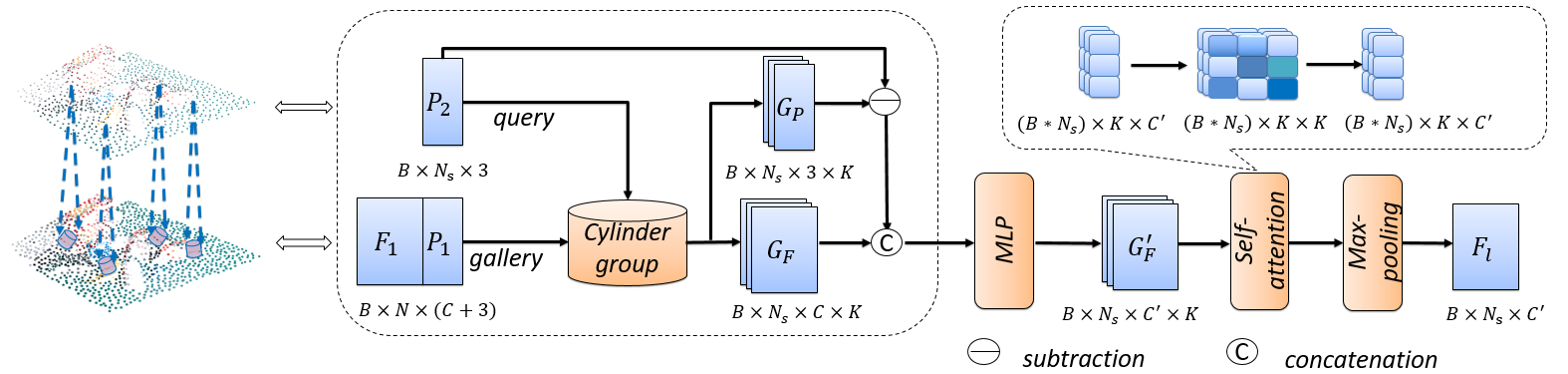}}
\end{minipage}
\caption{Local attention module. Left: the diagram shows the process of cylindrical region query between cross-resolution point clouds; Middle: the detailed architecture of cylinder grouping, $P_1$, $P_2$ are respectively high and low resolution point sets from PointNet++ backbone, $N_s$ is the number of seed points, $K$ is the sample number during cylinder grouping; Right: local attention operation outputs final features map $F_l$, the top part of self-attention unit indicates the dimension of attention map is $K\times K$.}
\label{fig:local}
\end{figure*}

\textbf{Hybrid Physical Metric}.
The hybrid physical metric is a combination of our proposed metrics above and previous force-closure metric. Since different physical viewpoints are adopted, our proposed metric possesses better generalization ability of grasping reasoning. The force-closure metric generates a ten-bin grasp confidence score $S_t$. The final grasp score $S$ is computed as follows.
\begin{equation}
\label{eq:score}
    S = \lambda_t \cdot S_t + \lambda_f\cdot S_f + \lambda_g\cdot S_g + \lambda_c\cdot S_c
\end{equation}
We set $\lambda_t, \lambda_f, \lambda_g, \lambda_c = 0.7, 0.2, 0.05, 0.05$ in practice.

\section{FGC-GRASPNET}

\subsection{Overview}
To detect grasp poses in scene-level point clouds, each grasp element constituting the grasp representation needs to be predicted by the network. According to \cite{fang2020graspnet}, the foreground segmentation needs to be carried out at first, then the grasp representation prediction is decoupled into multiple tasks including the prediction of depth, width, approaching vector and the in-plane rotation. In predecessor tasks, the foreground segmentation considers the overall geometry structure and the grasp approaching vector prediction is closely associated with the direction to the ground in the world space. In comparison, in post tasks, the in-plane rotation, depth and width are usually related to the local geometry of a single object. Therefore, we design a multi-resolution architecture to extract features for these multiple tasks. As is shown in \figref{fig:net}, our network consists of a base backbone PointNet++\cite{qi2017pointnet++} and two branches called FA-head and RD-head. PointNet++\cite{qi2017pointnet++} is leveraged to extract features of hierarchical point sets. Features of the low-resolution seed point set pass into the FA-head for the foreground segmentation and point-wise approach direction score regression, while features of high-resolution point set are served for post tasks in the RD-head through our designed local attention module. We finally adopt a new joint learning loss. On one hand, considering our finer confidence score, we perform the regression operation for all the predicted scores instead of only regressing their maximal values. In this way, the network can be supervised more sufficiently. On other hand, since the supervision information for the maximal predicted score is weakened, we add a loss for the depth prediction task. Our new designed loss can be better adapted to our hybrid physical metric.

\subsection{Local Attention Module}
The in-plane rotation and depth prediction usually depend on the local geometry structure of the single object model. To extract more abundant and complex local information, we design a local attention module. This module queries region information in high-resolution feature map and evolves it by the self-attention unit\cite{vaswani2017attention}. Its structure is shown in the \figref{fig:local}. First, cylinder region query is employed to search the neighbor points of the seed point set $P_2$ in the high-resolution point set $P_1$. More cylinder grouping details can be referred to\cite{fang2020graspnet}. This grouping process outputs point features $G_F \in \mathbb{R}^{B\times N_s \times C \times K}$ and point coordinates $G_P\in \mathbb{R}^{B\times N_s \times 3 \times K} $. Then, $G_F$ is concatenated with the coordinate offsets between query points $P_2$ and group points $G_P$. This concatenated feature will be processed by a self-attention layer to enhance the local region attention. The self-attention layer only focuses on capturing region-range contextual information, and its details can be referred to \cite{guo2020pct, zhao2020point}. These group features finally forward through the max pooling layer along the K-dimension to retain the most salient features. Our local attention module integrates the features and coordinates of spatial neighbor points, which better adapts to the in-plane rotation and depth classification tasks.

\begin{table*}[t]
\centering
\caption{evaluation for different models. The table shows the results on data captured by RealSense/Kinect respectively}
\label{table:map}
\begin{tabular}{|cc|ccc|ccc|ccc|}
\hline
\multicolumn{2}{|c|}{\multirow{2}{*}{Models}}& \multicolumn{3}{c|}{Seen}&\multicolumn{3}{c|}{Unseen}&\multicolumn{3}{c|}{Novel}\\
\cline{3-11}
\multicolumn{2}{|c|}{} & mAP & AP$_{0.3}$ & AP$_{0.7}$ 
& mAP & AP$_{0.3}$ & AP$_{0.7}$
& mAP & AP$_{0.3}$ & AP$_{0.7}$\\
\hline

\multicolumn{2}{|c|}{Fang et al.\cite{fang2020graspnet}}	&37.40/32.79	&34.75/27.79	&18.79/14.75	&35.01/30.45  &30.25/24.34	&17.54/11.18	&23.22/21.05	&12.36/9.29	 &3.21/2.48\\ 
\hline

\multicolumn{2}{|c|}{Ours(no depth)}&
41.88/37.53 &41.59/34.65 & 24.11/20.11 & 36.75/31.41 &33.47/25.66 & 19.21/12.18 & 24.48/21.81 & 14.61/10.22 &3.89/2.61\\
\hline
\multicolumn{2}{|c|}{Ours}&
49.68/41.09 &53.06/40.18 &	33.73/23.58 &40.09/33.05 &38.40/28.35 &23.31/13.64 &	26.01/23.27 &17.37/12.31 &5.03/3.35\\

\hline
\end{tabular}
\end{table*}
\begin{table*}[t]
\centering
\caption{Results of Single object grasping tests. FCM is force-closure metric, HPM is hybrid physical metric.}
\label{table:single}
\begin{tabular}{|cccc|cc|cc|cc|cc|cc|cc|cc|cc|cc|}
\hline

\multicolumn{4}{|c|}{Object/ID}& 
\multicolumn{2}{|c|}{Banana/5}& 
\multicolumn{2}{c|}{Apple/12}&
\multicolumn{2}{c|}{Nivea.../42}&
\multicolumn{2}{c|}{Hosjam/44}&
\multicolumn{2}{c|}{Giraffc/55}&
\multicolumn{2}{c|}{Weiquan/57}&
\multicolumn{2}{c|}{Darlic../58}&
\multicolumn{2}{c|}{Lion/67}&
\multicolumn{2}{c|}{All}\\
\hline

\multicolumn{4}{|c|}{Attempt}& 
\multicolumn{2}{|c|}{30}& 
\multicolumn{2}{c|}{30}&
\multicolumn{2}{c|}{30}&
\multicolumn{2}{c|}{30}&
\multicolumn{2}{c|}{30}&
\multicolumn{2}{c|}{30}&
\multicolumn{2}{c|}{30}&
\multicolumn{2}{c|}{30}&
\multicolumn{2}{c|}{240}\\
\hline

\multicolumn{2}{|c|}{\multirow{2}{*}{FCM}}&
\multicolumn{2}{|c|}{Success}& 
\multicolumn{2}{|c|}{25}& 
\multicolumn{2}{c|}{25}&
\multicolumn{2}{c|}{17}&
\multicolumn{2}{c|}{28}&
\multicolumn{2}{c|}{24}&
\multicolumn{2}{c|}{27}&
\multicolumn{2}{c|}{27}&
\multicolumn{2}{c|}{24}&
\multicolumn{2}{c|}{197}\\
\cline{3-22}

\multicolumn{2}{|c|}{}&
\multicolumn{2}{|c|}{Success Rate}& 
\multicolumn{2}{|c|}{$83.33\%$}& 
\multicolumn{2}{|c|}{$83.33\%$}& 
\multicolumn{2}{c|}{$56.67\%$}&
\multicolumn{2}{c|}{$93.33\%$}&
\multicolumn{2}{c|}{$80\%$}&
\multicolumn{2}{c|}{$90\%$}&
\multicolumn{2}{c|}{$90\%$}&
\multicolumn{2}{c|}{$80\%$}&
\multicolumn{2}{c|}{$82.08\%$}\\
\hline

\multicolumn{2}{|c|}{\multirow{2}{*}{HPM}}&
\multicolumn{2}{|c|}{Success}& 
\multicolumn{2}{|c|}{27}& 
\multicolumn{2}{c|}{25}&
\multicolumn{2}{c|}{22}&
\multicolumn{2}{c|}{29}&
\multicolumn{2}{c|}{26}&
\multicolumn{2}{c|}{28}&
\multicolumn{2}{c|}{27}&
\multicolumn{2}{c|}{25}&
\multicolumn{2}{c|}{209}\\
\cline{3-22}

\multicolumn{2}{|c|}{}&
\multicolumn{2}{|c|}{Success Rate}& 
\multicolumn{2}{|c|}{$90\%$}& 
\multicolumn{2}{|c|}{$83.33\%$}& 
\multicolumn{2}{c|}{$70\%$}&
\multicolumn{2}{c|}{$96.67\%$}&
\multicolumn{2}{c|}{$86.67\%$}&
\multicolumn{2}{c|}{$93.33\%$}&
\multicolumn{2}{c|}{$90\%$}&
\multicolumn{2}{c|}{$83.33\%$}&
\multicolumn{2}{c|}{$87.08\%$}\\
\hline

\end{tabular}
\end{table*}
\subsection{Loss function}
For the predecessor tasks in FA-head, we adopt a classification loss to learn the object mask, and a regression loss for view scores to supervise the approach direction learning. The loss $L_{pre}$ is as follow:
\begin{equation}
    L_{pre} = \frac{1}{N}\sum_{i=1}^{N}L_{cls}(\hat{m}_i, m_i) + \frac{1}{N_{reg}}\sum_{i=1}^{N}\sum_{j=1}^{V}m_i\cdot L_{reg}(\hat{s}_{ij}, s_{ij})
\end{equation}
where $m_i$ is a binary label that it is assigned 1 if the point is of objects, $s_{ij}$ is the view score label used the maximum grasp score in each approach direction, $\hat{m}_i$, $\hat{s}_{ij}$ represent the corresponding predicted values. $N=1024$ and $V=300$.

For the post tasks in RD-head, we regress all 48 grasp scores which are correspond to 48 grasp proposals composed of 12 types of rotation and 4 types of depth. In addition, we predict the rotation direction and depth through a classification loss. The grasp width is regressed along the prediction rotation category. The loss $L_{post}$ is formulated as follows: 
\begin{equation}
\begin{split}
    &L_{post} =\frac{1}{N}\sum_{i=1}^{N}
   \Big( \frac{1}{N_{reg}}\sum_{j=1}^{A\times D}m_i\cdot L_{reg}(\hat{s}_{ij}, s_{ij})\\
    &+ \frac{1}{N_{cls}}\sum_{j=1}^{A}m_i\cdot L_{cls}(\hat{r}_{ij}, r_{j})+
    \frac{1}{N_{cls}}\sum_{j=1}^{D}m_i\cdot L_{cls}(\hat{d}_{ij}, d_{j})\\
    &+\alpha\cdot\frac{1}{N_{reg}}\sum_{i=1}^{D}m_i\cdot L_{reg}(\hat{w}_{ij}, w_{ij})
   \Big)
\end{split}
\end{equation}
where $m_i$ is the same in $L_{pre}$, $s_{ij}$ is 48 candidate score labels on the top view direction predicted by the FA-head, $r_{i}$ is the index of maximum grasp score among 12 in-plane rotation directions and $d_{i}$ is the index of maximum grasp score among 4 depths, $w_{ij}$ is grasp width of the $r_{i}$ direction grasp pose, $\hat{s}_{ij}$, $\hat{r}_{ij}$, $\hat{d}_{ij}$, $\hat{w}_{ij}$ represent the corresponding prediction values, and $A=12$, $D=4$, we set $\alpha=0.2$.

Finally, the overall joint learning loss $L$ is;
\begin{equation}
    L = L_{pre}+\beta\cdot L_{post}
\end{equation}
We set $\beta=0.3$ in practice.

\section{EXPERIMENTS}

\subsection{Network Evaluation}
\subsubsection{Evaluation Metric}
We conduct experiments on the GraspNet-1Billion dataset\cite{fang2020graspnet}, and leverage our new grasp confidence scores according to \myref{eq:score}. We use the Average Precision (AP)\cite{fang2020graspnet} to evaluate the network. Specifically, the grasp pose non-maximum suppression (NMS) and collision detection are used to filter candidates at first. Then, we extract candidates with the top-50 predicted scores, and query the corresponding real confidence scores. We set up different score thresholds, then calculate the mean AP (mAP) under different score thresholds $[0, 0.1, 0.3, 0.5, 0.7, 0.9]$ , which are corresponding to friction coefficient thresholds in \cite{fang2020graspnet}.

\subsubsection{Comparison Studies}
We compare the performance of networks under the new confidence scores generated by the hybrid physical metric. Other parameter settings of the experiments are all the same.  Since the numerical distribution of the grasp confidence scores has changed, the AP results are different from \cite{fang2020graspnet}. As is illustrated in Table \ref{table:map}, our FGC-GraspNet outperforms the network of \cite{fang2020graspnet} on both Realsense camera dataset and Kinect camera dataset. On the seen test set, our FGC-GraspNet outperforms previous method by more than 10\% AP. On the unseen or novel test set, the improvement is also quite significant. This demonstrates the effectiveness of the FGC-GraspNet.

\subsubsection{Ablation Studies}
We also conduct ablation experiments to analysis the contributions of the depth classification loss in our new joint loss function. The results are listed in Table \ref{table:map}. Ours(no depth) indicates that we do not adopt the depth classification loss function in the joint learning loss. Instead, we predict the depth by using different height of cylinder like \cite{fang2020graspnet}. We notice that our depth loss generally improves the results by more than 7\% on the seen test set. This indicates that our network is greatly improved by the depth classification loss function. 

\begin{table}[t]
\begin{minipage}{0.5\textwidth}
\centering
\caption{Results of scene grasping.}
\label{table:success}
\begin{tabular}{|c|c|c|cc|cc|}
\hline

\multicolumn{1}{|c|}{\multirow{2}{*}{Scene}}& 
\multicolumn{1}{|c|}{\multirow{2}{*}{Object ID}}&
\multicolumn{1}{|c|}{\multirow{2}{*}{Attempt}}&
\multicolumn{4}{c|}{Success Rate}\\
\cline{4-7}
\multicolumn{1}{|c|}{}&\multicolumn{1}{|c|}{}&\multicolumn{1}{|c|}{}&\multicolumn{2}{|c|}{Fang et al.\cite{fang2020graspnet}}&\multicolumn{2}{c|}{Ours}\\
\hline

\multicolumn{1}{|c|}{Scene1}& 
\multicolumn{1}{c|}{5,12,39,44,67}&
\multicolumn{1}{c|}{10}&
\multicolumn{2}{c|}{90$\%$}&
\multicolumn{2}{c|}{92$\%$} \\
\hline

\multicolumn{1}{|c|}{Scene2}& 
\multicolumn{1}{c|}{30,42,53,59,61}&
\multicolumn{1}{c|}{10}&
\multicolumn{2}{c|}{82$\%$}&
\multicolumn{2}{c|}{86$\%$} \\
\hline

\multicolumn{1}{|c|}{Scene3}& 
\multicolumn{1}{c|}{35,36,55,57,63}&
\multicolumn{1}{c|}{10}&
\multicolumn{2}{c|}{92$\%$}&
\multicolumn{2}{c|}{92$\%$} \\
\hline

\multicolumn{1}{|c|}{Scene4}& 
\multicolumn{1}{c|}{37,38,40,48,68}&
\multicolumn{1}{c|}{10}&
\multicolumn{2}{c|}{84$\%$}&
\multicolumn{2}{c|}{92$\%$} \\
\hline

\multicolumn{1}{|c|}{All}& 
\multicolumn{1}{c|}{All}&
\multicolumn{1}{c|}{40}&
\multicolumn{2}{c|}{87$\%$}&
\multicolumn{2}{c|}{90.5$\%$} \\
\hline

\end{tabular}
\end{minipage}
\end{table}

\begin{figure*}[t]
\centering
\begin{minipage}{\textwidth}
\center{\includegraphics[width=\linewidth]{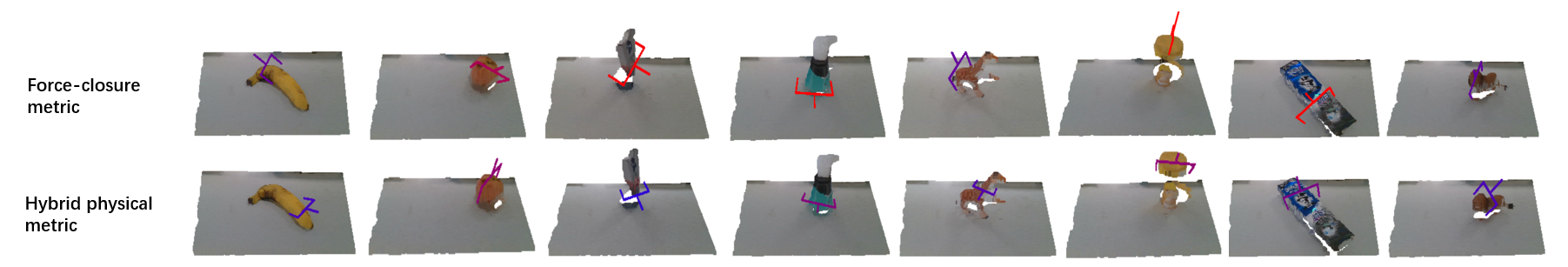}}
\end{minipage}
\caption{Single object grasping. Each picture show the top score grasp candidate predicted by using corresponding confidence scores generated by force-closure metric or hybrid physical metric. The order of objects is from banana to lion, which is consistent with Table \ref{table:single}.}
\label{fig:single}
\end{figure*}

\begin{figure*}[t]
\centering
\begin{minipage}{\textwidth}
\center{\includegraphics[width=\linewidth]{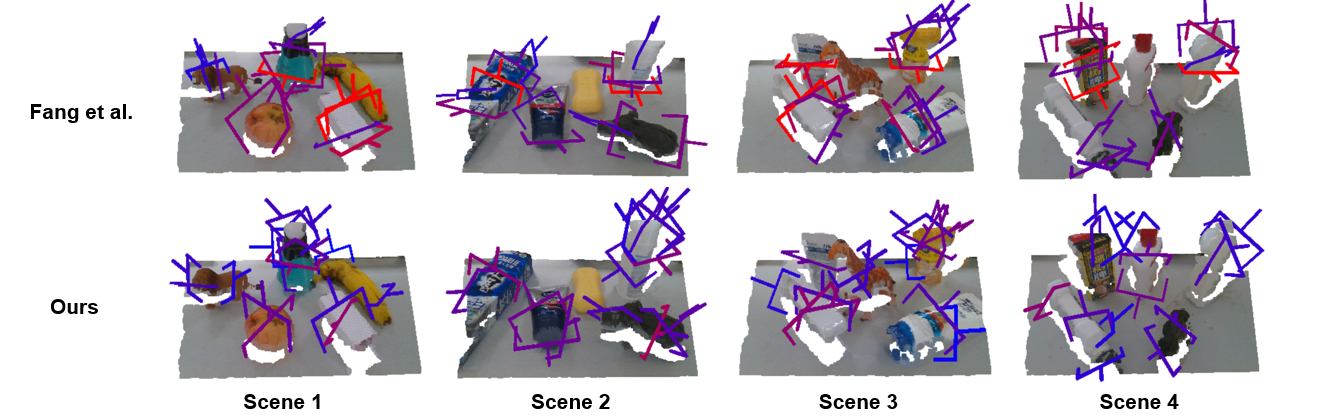}}
\end{minipage}
\caption{Scene grasping. There are 15 top grasp candidates after NMS in each picture. The red color represents higher prediction confidence score and the color contrast can only be reflected in the same model. These scenes are consistent with Table \ref{table:success}. }
\label{fig:scene}
\end{figure*}

\subsection{Hybrid Physical Metric Evaluation}
To demonstrate the efficiency of our hybrid physical metric, we conduct single object grasping experiments in the real robotic environment. We select 8 objects in tests. To ensure the robustness of the experiment, the geometric shapes of these objects are representative. We use the FGC-GraspNet to obtain two models. One is trained in new grasp scores generated from the hybrid physical metric, and another is trained in original score from the force-closure metric. We perform 30 grasp attempts for each object and count the successful grasps. We change the position and pose of the object every time to increase the reliability of the test. The results are reported in Table \ref{table:single}. We observe that the introduction of hybrid metrics improves the success rate of grasping in the real world. We show some examples of the compared results under two metrics in the \figref{fig:single}. We can see that the prediction grasps under hybrid physical metric are prone to show the following characteristics: searching the flat contact point, keeping the antipodal direction consistent with the normal direction of the flat contact point, closing to the center of the object, and avoiding too close between gripper end points and contact points.

\subsection{Real Robotic Experiment}
The robotic experimental setup includes a UR3 robotic arm and an OnRobot RG2 gripper. The computational  resources include NVIDIA TITAN Xp GPU and Intel Xeon E5-2650 CPU. We collect RGB and depth images from an Intel Realsense D435i camera. We set up 4 cluttered table scenes with 5 different objects and conduct several grasp experiments in these scenes. The target of these experiments is to move the objects to the storage box. The camera keeps the same pose in a complete test when collecting the scene images. After predicting grasp results, the top score grasp pose will be given to the robotic arm. It will be regarded as a successful grasp if the object is and placed in the storage box. We set each attempt of one scene for 5 times of image collection and robotic arm motion operation. Therefore, the single attempt success rate is calculated as the percentage of successful grasps among 5 times. Then we make 10 attempts for each scene and obtain the average success rate. We test \cite{fang2020graspnet} and our model respectively. Partial visualization results under four scenes are shown in \figref{fig:scene}. The success rate results are reported in Table \ref{table:success}. The experiment results prove that our method improves the performance of grasping in the real world.

\section{CONCLUSIONS}
We present the hybrid physical metric to evaluate the grasp quality for 6-DoF grasp pose detection. We conduct this metric to generate new grasp confidence scores on the GraspNet-1Billion dataset. We further propose a multi-resolution network FGC-GraspNet to learn these confidence scores better. Altogether, the network evaluation and adequate real robot experiments show that both the hybrid physical metric and the FGC-GraspNet play a positive effect on improving the success rate of grasping. In future work, we aim to apply this framework into complex integrated robotic tasks like feeding a person, cooking or table cleaning.

\section{ACKNOWLEDGMENT}
This work was supported by the state key development program in 14th Five-Year under Grant No.2021YFF0602103, 021YFF0602102, 2021QY1702. We also thank for the research fund under Grant No.2019GQG0001 from the Institute for Guo Qiang,Tsinghua University.

\addtolength{\textheight}{0cm}   





\bibliographystyle{plain}
\bibliography{ref}

\end{document}